# Character Feature Engineering for Japanese Word Segmentation

**Mike Tian-Jian Jiang**
Independent[†]
tmjiang@gmail.com

## Abstract

On word segmentation problem, machine learning architecture engineering often draws attention. The problem representation itself, however, has remained almost static as either word lattice ranking or character sequence tagging, for at least two decades. The latter often shows stronger predictive power than the former for out-of-vocabulary (OOV) issue. When the issue escalating to rapid adaptation, which is a common scenario for industrial applications, active learning of partial annotations or re-training with additional lexical resources is usually applied, however, from a somewhat word-based perspective. Not only it is uneasy for end users to comply linguistically consistent word boundary decisions, but also the risk/cost of forking models permanently with estimated weights is seldom affordable. To overcome the obstacle, this work provides an alternative, which uses linguistic intuition about character compositions, such that a sophisticated feature set and its derived scheme can enable dynamic lexicon expansion with the model remaining intact. Experiment results suggest that the proposed solution, with or without external lexemes, performs competitively in terms of F1 score and OOV recall across various datasets.

## 1 Introduction

According to ISO/DIS 24614-1, word segmentation is a process to divide a sentence into meaningful tokens called "word" conventionally (Choi et al., 2009). This process is usually considered fundamental and essential for many Asian languages to properly deal with downstream natural language processing applications. Unlike most writing systems in the world, an Asian language like Japanese normally retain no specific symbols such as whitespace for being word boundary delimiter, and word boundaries are often ambiguous if only looking up lexemes without taking context into account. Furthermore, Japanese writing system comprises three types of scripts, namely hiragana, katakana, and Chinese character in Sino-Japanese form (referred as kanji hereafter) (Joyce et al., 2012). For example, considering a phrase in gibberish "Lubbadubdub!" to be segmented as "Lubba dub-dub !" whereas "Lu" can be a word and "bad" can be another under different circumstances, while "L" could be "l" or "ℓ" in other types of scripts. Even when representing the same meaning, with or without a hyphen indicating a compound can formulate alternate standards of word segmentation. On the other hand, once the inevitable out-of-vocabulary (OOV) situation occurs with diverse language varieties, genres, registers, or domains, a robust Asian word segmentation system is expected to induce and adapt unseen usages morphologically (Tseng et al., 2005; Saito et al., 2014; Morita et al., 2015; Xu et al., 2006; Li et al., 2015; Jin and Wong, 2002; Gao and Stephan, 2010; Murawaki and Kurohashi, 2010), say realizing "Wubba" being an unknown word and then correctly segment another sentence in gibberish "Wubba lubba dub dub !"

Japanese word segmentation (JWS) task has been mostly integrated within morphological analysis (MA) task, which not only splits an input sentence into words, but also jointly annotates morphemes with their corresponding part-of-speech (POS) tags. The joint learning task has usually been defined as a word lattice ranking problem and approached differently with handcrafted rules[1], hidden Markov models (HMMs) (Nagata, 1994; Asahara and Matsumoto, 2000), maximum entropy Markov models (MEMMs) (Uchimoto et al., 2001; Uchimoto et al., 2002; Uchimoto et al., 2003), support vector machines (SVMs) (Sassano, 2002), linear-chain conditional random fields (linear-chain CRFs) (Kudo et al., 2004), averaged structured perceptron (Kaji and Kitsuregawa, 2013), exact soft confidence-weighted learning with recurrent neural network language model (RNNLM) estimated words (Morita et al., 2015), etc. With the intention of exploring a wider range of approaches, Chinese word segmentation (CWS) works may be relevant. Because not only kanji is believed being a core building block for Japanese morphology (Joyce et al., 2014), but also affecting adjacent hiragana/katakana morpho-phonologically (Irwin, 2005; Kurisu, 2000; mis,; Kawahara and Nishimura, 2002; mis, 1998; mis,; mis, 1996). So far the dominant viewpoint for CWS task, however, is character sequence tagging (Huang and Hai, 2007; Hai et al., 2017), and (Ng and Low, 2004) show that POS-joint learning might be optional. Intriguingly, recent developments on JWS emerge to character-based methods with POS (Nakagawa and Uchimoto, 2007) or without it, either performing a two-step MA (Neubig et al., 2011) or just JWS itself (Nakagawa, 2004; Kitagawa and Komachi, 2017), while some of CWS researches begin revisiting word-based (Cai and Zhao, 2016; Cai et al., 2017) ones.

---

[†] This work was partially done while the author was affiliated with DG Lab, Digital Garage, Inc.
[1] http://nlp.ist.i.kyoto-u.ac.jp/EN/index.php?JUMAN

This work then aims to empirically deepen the understanding of character compositionality on JWS. The expected contribution is twofold: first rationalizing a systematic label and feature induction procedure, and secondly utilizing the outcome to demonstrate dynamic lexicon expansion in a pragmatic fashion. Although a previous work have demonstrated that active learning with partially annotated keyword-in-context (KWIC) is more effective than lexicon expansion (Mori and Neubig, 2014), KWIC acquisition can be still costly for rapid adaptation. Despite partial annotations are relatively easier to acquire than thoroughly curated corpora, it is likely that industrial/end-users possess no linguistic expertise but domain knowledge. In this work, the hypothesis is that one can form each user-defined word dynamically, by properly engineering innate variables of character sequence, instead of having the original model forked permanently with an estimated weight for every unknown lexeme.

## 2 Reproducibility

### 2.1 Evaluation Metrics, Significant Figure, and Statistical Significance Test

This work follows the convention of JWS, CWS, and many other natural language processing (NLP) tasks, using $F_1$ score in terms of character and word. Some works treat them based on word boundary or longest common subsequence. This work opts out of those treatments since they are virtually correlated to the character/word based ones. As for the $F_1$ scores in percentage, the second decimal place sometimes suffers from randomness according to preliminary tests, especially when experiments are conducted with inconsistent machine/compiler combinations. Despite character-wise scores being reported with two decimal places for the purpose of discussion, the significant figure should be always the first one. The significant figure difference may also be related to the character-word proportions, which can be somewhat deduced from Table 1 in the next subsection of datasets. For the choice of evaluation metrics itself, unfortunately there are several studies reporting incomparability issues, due to the differences ranging from downstream application requirements (Jiang et al., 2011), cognitive costs (Qian et al., 2016), segmentation standard disagreements (Shao et al., 2017), to the inherent Prevalence/Bias inconsistencies among various samples and systems (Powers, 2011), in spite of morphophonemics for Japanese and other languages usually apply unbiased metrics such as Markedness (Irwin, 2005). Ironically, while dataset-oriented bias will be an issue addressed in the next subsection, the point of this work can be seen as fitting it as much as possible with extensive feature engineering. Nevertheless, the biased metrics such as $F_1$ score will likely render its statistical significance tests questionable among non-equivalent works even with the same dataset. Sometimes sequential labeling tasks in general adopt confidence intervals (Sproat and Emerson, 2003; Emerson, 2005; Levow, 2006; Jin and Chen, 2008) and McNemar's test (Sha and Pereira, 2003; Kudo et al., 2004; Song and Sarkar, 2008; Shen et al., 2014; Matsumoto et al., 2004; Okanohara et al., 2006; Fujinuma and Grissom, 2017), whereas both of them are arguably too conservative (Fagerland et al., 2013; Wolfe and Hanley, 2002; Goldstein and Healy, 1995), and the majority of JWS/CWS works seem happy without them. Proper significance tests may exist, e.g., the mid-p variation of McNemar's test (Fagerland et al., 2013), yet this work would like to stay ignorant for the time being, unless their accessibilities to JWS has been further established.

| Reference | Doc. Type | #Sentence | | Spec. | | #Word | | #IVs | #Character | |
|---|---|---|---|---|---|---|---|---|---|---|
| | | Training | Test | T | UW | Training | Test | | Training | Test |
| This work | HOM | 56,760 | 3,010 | O | S | 1,203,331 | 67,435 | 45,477 | 1,906,452 | 105,323 |
| | | | | | SS | 1,323,653 | 74,054 | 40,995 | | |
| | | | | N | S | 1,196,233 | 67,089 | 45,477 | 1,908,733 | 105,491 |
| | | | | | SS | 1,316,555 | 73,708 | 40,977 | | |
| K. & M. '17 | | 56,448 | 2,984 | N | SS | - | - | - | - | - |
| NextNLP-MA | | 57,281 | 3,024 | | | - | 74,865 | - | - | 106,661 |
| This work | GEN | 31,064 | 2,091 | N | SS | 816,882 | 64,727 | 28,733 | 1,180,008 | 92,642 |
| | HET | 5,720 | 658 | | | 115,070 | 12,091 | 31,305 | 159,159 | 16,838 |
| M. & N. '14 | GEN | - | - | | | 784k | - | 29.7k | - | - |
| | HET | - | - | | | 114k | 13.0k | 32.5k | - | - |
| M. et al. '11 | GEN | 27,338 | 3,038 | | | 782,584 | 87,458 | 28,315 | 1,131,317 | 126,154 |
| | HET | 5,800 | 645 | | | 114,265 | 13,018 | - | 158,000 | 17,980 |
| N. et al. '11 | GEN | - | - | | | 782k | 87.5k | - | - | - |
| | HET | - | - | | | 153k | 17.3k | - | - | - |

Table 1: Comparison of Datasets Arranged from BCCWJ.
Each cell containing only a '-' indicates the number is unreported from the reference.

## 2.2 Datasets

This work studies JWS with version 1.1 of the Balanced Corpus of Contemporary Written Japanese (BCCWJ) (Maekawa et al., 2013) that contains modern texts written in multiple domains and registers. For the sake of reproducibility and with respect to strictly closed test criteria of machine learning common practice, sub-subsections below describe dataset specifications that are compatible with previous works.

**Construction:** When both analyzing overall phenomena and investigating adaptation ability, the document file ID list[2] applied for extracting test set here, is identical to what has been assigned by the MA team of Project Next NLP (NextNLP-MA)[3]. This work arbitrarily picks IDs from the training set to define a reusable development set[4] for hyper-parameter tuning. For adaptation, although it is preferred to have a domain/register-specific separation setup like what a series of JWS works defined (Mori et al., 2011a; Mori et al., 2011b; Neubig et al., 2011; Mori and Neubig, 2014), it is unfortunately uneasy to reproduce. They select "Yahoo! Answers" documents as web texts for adaptation target. For the target's counterpart, some works see book, news, and whitepaper files as generic texts (Mori et al., 2011a; Mori et al., 2011b; Neubig et al., 2011), while another one additionally include Yahoo! Blog and magazine files (Mori and Neubig, 2014). According to yet another study (Mori et al., 2011b), each remainder of a document ID's serial number divided by 10 is used to partition training/test sets for both web texts and generic texts, which is the only known description to regenerate the data sets, but this work still fails to replicate them with acceptable margin of word counts.

**Script:** The latest BCCWJ provides two script variations across the whole corpus, namely original texts (OT) and their number-transformed (NT) counterparts, where consecutive digits and separators are translated into corresponding Han-character numbers.

**Granularity:** BCCWJ regulation defines rules to form morphemes into short-unit word (SUW). (Mori et al., 2014) derive it with inflectional verbs further split into stems and suffixes, hence super-short-unit word (SSUW), which can be prepared by NextNLP-MA's conversion tool[5] with a patch[6] to accommodate the latest BCCWJ.

**Comparison:** Table 1 lists statistics and traits of dataset constructs for this and related works, in order to examine how comparable the results among them would be. For related works that have used BCCWJ, "K. & M. '17" refers to one of the latest JWS work (Kitagawa and Komachi, 2017), "M. & N. '14" denotes a study of language resource addition (Mori and Neubig, 2014), while "M. et al. '11" covers two related works (Mori et al., 2011a; Mori et al., 2011b), and "N. et al. '11" stands for another (Neubig et al., 2011). The column "Doc. Type" indicates whether the dataset is for adaptation to web texts that are heterogeneous (HET) to the generic ones (GEN), or just for typically homogeneous training/test sets (HOM) designated by NextNLP-MA's file ID list. Among the datasets of the HOM type, NextNLP-MA's statistics seems unmatched with published works except the website and its listed slides. Besides, sentence counts of this type all vary a little, possibly due to the version differences of BCCWJ. For script and granularity specifications, it is as expected that, word counts between OT and NT groups are slightly different, and character counts remain intact within each group, since the SSUW treatment is fully reversible. A much bigger concern lies in the GEN-HET partitioned sets, where count/ratio gaps exist between this and related works, especially for GEN parts.

---
[2] http://plata.ar.media.kyoto-u.ac.jp/mori/research/NLR/JDC/ClassA-1.list
[3] http://plata.ar.media.kyoto-u.ac.jp/mori/research/topics/PST/NextNLP.html
[4] Anonymous for the time being
[5] http://plata.ar.media.kyoto-u.ac.jp/mori/research/NLR/JDC/bccwjconv.tar.gz
[6] An anonymous patch file for the time being

| Labels | Label Strings by Word Length n | |
|---|---|---|
| | 1 | > 1 |
| B, I | B | BI{1, n-1} |
| B, I, S | S | BI{1, n-1} |
| B, I, E | B | BE; BI{1, n-2}E |
| B, I, E, S | S | BE; BI{1, n-2}E |
| B, 2, I, E, S | S | BE; B2E; B2I{1, n-3}E |
| B, 2, 3, I, E, S | S | BE; B2E; B23E; B23I{1, n-4}E |
| … | S | … |

Table 2: Label Variations

| C | O | L | T | R | $L_R$ | S | $L_S$ |
|---|---|---|---|---|---|---|---|
| L | 0 | 5 | $L$ | Lu | 2 | Lu | 2 |
| u | 1 | | | dub, Lu | 2, 3 | Lu | 2 |
| b | 2 | | | bad, dub | 3 | bad | 3 |
| b | 3 | | | | | | |
| a | 4 | | | a, bad | 2 | a, bad | 1, 3 |
| d | 0 | 3 | | bad, dub | 3 | bad, dub | 3 |
| u | 1 | | | dub, Lu | 2, 3 | dub | 3 |
| b | 2 | | | bad, dub | 3 | dub | 3 |
| ! | 0 | 1 | $P$ | ! | 1 | ! | 1 |

Table 3: Sentence-wise Information

## 3 Character Sequence Tagging

Natural language processing often involves breaking a given string into smaller building blocks. After the inside-outside-beginning (IOB) tagging scheme has been introduced by a pioneer work (Ramashaw and Marcus, 1995), its variations studied for noun phrase (NP) chunking (Sang and Veenstra, 1999), and latter widely adopted by the SIGNLL Conference on Computational Natural Language Learning (CoNLL) for several tasks such as NP bracketing, phrase chunking, clause identification, named entity recognition, etc., therefore empirically proving this boundary-indicator tagging scheme is quite flexible. The resemblance between those tasks and word segmentation has probably inspired some participants of the Special Interest Group on Chinese Language Processing (SIGHAN) word segmentation bakeoffs. CWS-specific versions that classify Chinese characters for their word-belonging positions have gradually evolved ever since. A summary has examined those versions and drawn a conclusion that, while most recent works usually stick to a four-tag—BIES (or BMES) for beginning, inside (or middle), ending, and single character of a word, respectively—format, a six-tag extension with additional labels for the second and the third characters, is the best choice for contemporary Mandarin Chinese (Zhao et al., 2010). JWS-related works to date, however, may have only applied BIES (Kitagawa and Komachi, 2017) or two-tag (Neubig et al., 2011; Sassano, 2002) schemes.

Subsequently, this work intends to explore more complex labeling schemes. Table 2 demonstrates how those positional label variations extend forwardly. By producing combinations of class labels as random variables and feature tags as observed variables, many configurations are experimented with linear-chain CRFs (Lafferty et al., 2001), and their critical outcomes will be reported accordingly. As for hyper-parameter, $\ell^1$ and $\ell^2$ norms are roughly tuned to 0.000015 and 0.0025, respectively, based on the designated development set, and then applied to every experiment throughout.

### 3.1 Sentence-wise Information

Table 3 summarizes sentence-wise information about a test case in gibberish "Lubba dub !" and known-segments from an imaginary training set is just {a, bad, dub, Lu, !}, such that the test sentence suffers from OOV issues. For each character **C**, **O** indicates of-word offset and implies word boundaries; **L** records word length; **T** stands for character type based on Unicode character categories, e.g., *L* for letter and *P* for punctuation; **R** collects character appearances of any known segment; **S** selects **R** items that match substrings of the sentence; $L_R$ and $L_S$ count distinct lengths of **R** and **S** items, respectively. Real cases of course have more concrete pieces of information and their details will be discussed in latter subsections. **R** reflects the prior knowledge of lexicon. Besides that, **T** is the only external information of character mimicking practical situation. Information acquired by unsupervised methods, such as accessor variety (Feng et al., 2004), branching entropy (Jin and Tanaka-Ishii, 2006), frequent patterns (Jiang et al., 2012), etc., can be quite useful, but it is excluded to keep this work focused on character-word relations.

| Length Range | SUW | | SSUW | |
|---|---|---|---|---|
| | NT | OT | NT | OT |
| 1 | 56.01 | 57.05 | 66.70 | 67.59 |
| ≤ 2 | 90.86 | 90.91 | 93.50 | 93.53 |
| ≤ 3 | 96.19 | 96.21 | 97.07 | 97.08 |
| ≤ 4 | 98.71 | 98.71 | 98.94 | 98.94 |
| ≤ 5 | 99.45 | 99.45 | 99.52 | 99.52 |
| ≤ 6 | 99.78 | 99.79 | 99.81 | 99.81 |
| ≤ 7 | 99.92 | 99.92 | 99.92 | 99.92 |
| ≤ 8 | 99.96 | 99.96 | 99.96 | 99.96 |
| ≤ 9 | 99.97 | 99.97 | 99.98 | 99.98 |
| ≤ 10 | 99.98 | 99.98 | 99.98 | 99.98 |

Table 5: Word-Length Coverage (%)

| | BIES | | B23IES | |
|---|---|---|---|---|
| Word | | $F_1$ | | $F_1$ |
| | | 98.1 | | 98.3 |
| Label | #Correct | $F_1$ | #Correct | $F_1$ |
| B | 24,086 | 98.14 | 24,146 | 98.34 |
| 2 | - | - | 3,764 | 94.82 |
| 3 | - | - | 1,603 | 93.58 |
| I | 6,857 | 94.37 | 1,417 | 91.30 |
| E | 23,977 | 97.70 | 24,048 | 98.15 |
| S | 48,555 | 98.81 | 48,595 | 98.87 |
| All | 103,475 | 98.09 | 103,743 | 98.18 |

Table 4: BIES v. B23IES

Word-based methods utilize **R** directly as path-finding lattice with count-based weights, where $L_R$ implies implicitly. Since **R** is uneasy for character-based methods to incorporate explicitly, a feature type called "word indicator" (**W**) often identifies **S** with a limitation of maximum $L_S$, ranging from three to six characters. Some works even extend the word indicator to adjacent substrings, such that it imitates word bigrams (Sun, 2011; Sun, 2010). **T** has had been somewhat forbidden from strictly closed test for CWS bakeoffs (Sproat and Emerson, 2003; Emerson, 2005; Levow, 2006; Jin and Chen, 2008) before 2010 (mis,; mis, 2012; Duan et al., 2014), but still widely adopted anyway. In JWS, it is only natural to include **T**, considering Japanese writing system consists of three scripts, hiragana, katakana, and kanji. Generally **O** is used by character-based method to design class labels, yet its traits is embedded in **W**. What this work intends to explore are:

- Classify characters with not only **O** but also **L** as long as **T**;
- Invent a new feature type for $L_R$;
- Enhance **W** with **C**, **S** and a more general $L_S$.

### 3.2 Class Labels

A series of CWS works have reasoned that **O**/**L** coverage of a labeling scheme is related to its word segmentation performance, and concluded B23IES works best (Zhao et al., 2006; Zhao et al., 2010; Zhao and Kit, 2007; Huang and Zhao, 2006). The same method is hereby applied on BCCWJ and the result is listed in Table 5. The percentages suggest that B23IES can also work well on BCCWJ. To verify it, two vanilla models of linear-chain CRFs models for BIES and B23IES are trained with NT-SSUW dataset, using CRFsuite[7]. Features are just an intersection among common choices of previous works, namely character unigrams, bigrams, and (1-chararacter-jumped) skip-grams within a ±1-character window. The result shown in Table 4 confirms that B23IES is slightly better than BIES in terms of word $F_1$ score and recall of OOV ($R_{OOV}$). In terms of label-wise statistics, however, per class performance varies. Specifically, I of BIES classifies better than 2, 3, I of B23IES combined (6,857 > 6,784 = 3,764 + 1,603 + 1,417). Characters labeled with the same class may deserve to be more equal.

**Inspection:** Like the empirical supports for B23IES scheme in CWS, resemble linguistic intuitions may also stand for JWS (Joyce et al., 2014; Joyce et al., 2012; Ando and Lee, 2002; Fujinuma and Grissom, 2017), such as the second character of a three-character word may act differently than the one of a four-or-five-character word, cf. 2 for B23IES and I for BIES. However, judging by the earlier mentioned inferior performance of I for B23IES, the marginal probability $P_{B23IES}(I|B,2,3)$ could still be less appropriate than $P_{BIES}(I|B)$, even though label and length biases are supposedly light in linear-chain CRFs (Kudo et al., 2004). A negative ripple effect can also back-propagate for $P_{B23IES}(E|I)$ in comparison with $P_{BIES}(E|I)$. Additionally, perhaps character n-gram features within 3-character window just lack sufficient information. For instance, all of the above speculated issues might inhabit in a frequently used phrase, やりかた ‹ya ri ka-ta› "a way of doing," which consists of a noun かた ‹ka-ta› "method" (a. k. a 方) and a verb やる ‹ya-ru› "act" lexicalized with a continuative suffix り ‹ri›. B23IES and BIES models have concatenated it incorrectly as a four-character word and a pair of two-character words, respectively. On one hand, the correct segmentation is unseen in the training set, except for one of its equivalent forms やり方. On the other hand, 44 cases of four-character words begin with a consecutive やり, such as やりとり ‹ya-ri-to-ri› "give-and-take" since it is a compound noun. Those compounds have likely motivated B23IES model in favor of four-character words for やり being a disyllabic prefix. Meanwhile, 515 two-character words that contain monosyllabic や as a prefix or り as a suffix might just cause BIES model biased naively.

|  | **BIES** | **B23IES** | **BIES-LT$_2$** | **B23IES-LT$_2$** | **BIES-LT** | **B23IES-LT** |
|---|---|---|---|---|---|---|
| $\tau(y_{-1}x_{-1}x_0x_1, y_0)$ | 0.99665 | 0.99683 | 0.99677 | 0.99686 | 0.99527 | 0.99533 |
| #Class | 4 | 6 | 14 | 22 | 102 | 131 |

Table 6: G. & K.'s $\tau$ between Centered Character Trigram and Class Labels

**Correlation:** Goodman and Kruskal's tau (G. & K.'s $\tau$) tests have been applied in the preliminary study. G. & K.'s $\tau$ is convenient yet informative about conditional variable importance of machine learning (Strobl et al., 2008), which could provide some insight before blindly jumping into various models. With an R package[8], associations are measured between the centered character trigram $x_{-1}x_0x_1$ ($x_0$ indicates the character in question) and various label schemes. Besides the well-studied BIES and B23IES schemes, combinations of them and **L**/**T** information are tested, too. Interestingly, the trade-off between variability and sparseness is probably balanced when expanding BIES/B23IES with bi-class tags of both **L** and **T** (marked with a suffix **-LT$_2$**). Instead of multiplying the full populations of **L** and **T** (marked with a suffix **-LT**) and producing more than a hundred classes, bi-class **L** and **T** distinguish lengthy words and non-Japanese letters, respectively. Their specifications will be addressed in latter subsections. An important trick here is appending the trigram with the previous label $y_{-1}$ for simulating the edge feature of the first-order linear chain CRFs. The measurements listed in Table 6 may provide information for designing labels and features in addition to the word-length coverage from previous works, and become guidance in the next subsection. Other character n-grams are also tested and collaborated the same trend. Their statistics is skipped here for brevity.

### 3.3 Features

While G. & K.'s $\tau$ from various character n-grams to label schemes remain unlisted, an intriguing asymmetry should be noted here. Character n-grams using left-side characters associate stronger than their counterparts with right-side ones, which may correlate to the left-to-right fashion of the labeling scheme or of the underlying natural language behavior, perhaps both. Previous works of point-wise models choose feature n-grams in this way, partly due to the their boundary decision point is at the end of each word (Neubig et al., 2011; Sassano, 2002). The asymmetric importance of n-grams shown here could be one of explanations, and the next question would be if it still applies with structured models. A series of simple models for each n-gram variation in a ±2-character window are therefore built with liner-chain CRFs. Their $F_1$ scores for both in-vocabulary (IV) and OOV words are separately illustrated in Figure 2 and Figure 1, with horizontal axis denotes u, b, j, and t for unigram, bigram, skip-gram (jump), and trigram, respectively. They confirm not only the asymmetric tendency and a related work that has performed similar evaluations for feature induction (Ren and Li, 2017). The only notable exception here is for OOV $F_1$ between trigram's conjunctive forms, namely $x_{-2}x_{-1}x_0$ and $x_0x_1x_2$.

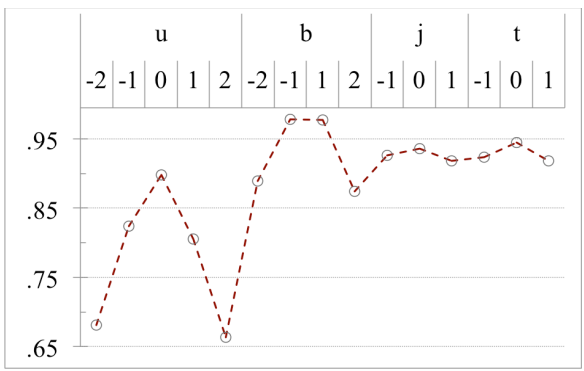

Figure 2. Individual Character n-gram's $F_{1\text{-IV}}$

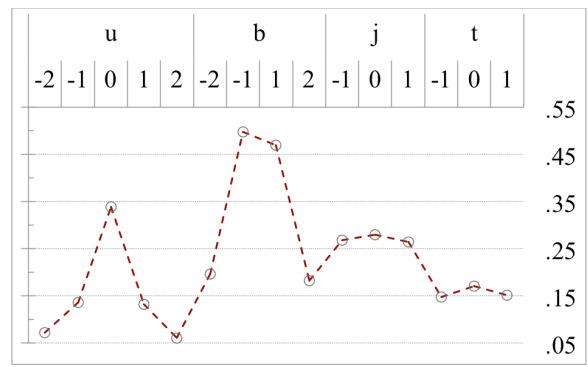

Figure 1. Individual Character n-gram's $F_{1\text{-OOV}}$

---
[8] https://cran.r-project.org/package=GoodmanKruskal

**Scaling:** An heuristic scaling trick has been implicitly applied first in a CWS work (Jiang et al., 2008) and then consciously reproduced (Wang et al., 2010b; Wang et al., 2010a) for its remarkable usefulness. The trick performs weight boosting by 2 for $x_0$ and by 3 for both $x_{-1}x_0$ and $x_0x_1$ of character n-gram features. The usefulness implies that the centered unigram and bigram characters have more predictive powers than other conjunctive ones. Their influences seem quantifiable if considering $F_1$ scores of IV and OOV at the same time. Incrementally testing feature conjunctions is a well-known technique for CRFs, and subsequently binning feature clusters by frequency scales is also a popular choice (McCallum, 2003; McCallum and Li, 2003; Peng et al., 2004; Peng and McCallum, 2004). On the other hand, studying n-gram behaviors for back-off in language modeling has a long tradition. A caveat is that regularization plays a crucial part for using back-off features, otherwise large weights may polarize the model (Sutton and McCallum, 2012). Besides regularizing and selecting by $\ell^1$ norm (Lavergne et al., 2010), normalizing real valued features to have zero-mean and unit-variance, a. k. a. feature standardization, is also a common technique. This work tries to exploit all the above experience as much as possible without information beyond sentence level, such that lexicon expansion on the fly can execute with a constant model. After evaluating feature conjunctions with certain metrics for IV and OOV words defined by the development set, a linear interpolation is then applied to the paired scores, i.e.: $(1-\alpha) \times$ $Score(\text{IV}) + \alpha \times Score(\text{OOV})$ for $0 \le \alpha \le 1$.

Finally, performing feature standardization to weigh each feature. In this work, a better choice of $Score(\cdot)$ is Recall rather than $F_1$, which may be task-oriented (Powers, 2011), and $\alpha$ is empirically set to 0.325 as a newly invented hyper-parameter.

**Length-category code (LC):** Utilizing the referred word length ($L_R$) and Unicode category (**T**) for each character is the next step in the plan. Preliminary test results help realize a new type of compound feature, which concatenate **T** and $L_R$ together as one. For example, the character 東 ‹tō› "east" has a "K|1234+" coding "K" for kanji character and "1234+" for that this character has participated in forming all-length words except five-character ones, where "+" represents for words longer than five characters.

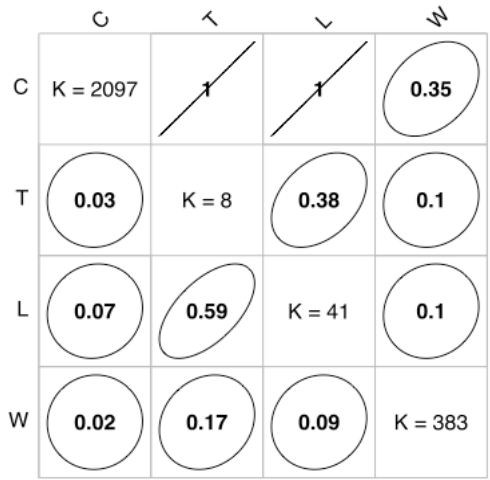

Figure 3: G. & K.'s $\tau$ Matrix of Sentence-wise Information

|  |  | $F_1$-Before | $F_1$-After |
|---|---|---|---|
| SSUW | NT | 98.7 | 98.9 |
|  | OT | 98.7 | 98.9 |
| SUW | NT | 98.7 | 99.0 |
|  | OT | 98.7 | 99.1 |

Table 8: Final Design's Performances Before/After Dynamic Lexicon Expansion

|  |  | $F_1$ |
|---|---|---|
| MeCab 0.996 | SUW-NT | 99.4 |
| KyTea 0.46 | SSUW-NT | 98.5 |
| KyTea 0.47 |  | 99.9 |

Table 7: Industrial-grade Accuracy

The rationale behind this compound feature is again based on G. & K.'s $\tau$. Figure 3 represents predictive powers between sentence-wise information, and the top two are $\tau(\mathbf{L_R}, \mathbf{T}) = 0.59$ and $\tau(\mathbf{T}, \mathbf{L_R}) = 0.38$. Separated and combined performances of them are evaluated like the character n-grams with the same scaling trick involved, and the outcome concurs.

**Word-character code (WC):** The last type of feature combines the in-place character (**C**), word-matched substring (**S**), and substring lengths (**L_S**). A vector of the latter two is usually called word indicator (**W**), and often implemented in one-hot encoding. In order to gain a better comprehension about the details, please refer to a previous work for a nicely drawn figure of it (Kitagawa and Komachi, 2017). This work relaxes the length limitation: instead of giving up on matching long words, the new implementation here will try to find all possible words, and encode any one longer than five characters into a preserved bucket as another code point, to match both the "+" treatment of length-category code and the coverage of B23IES labeling scheme. For example, when 東 appears next to 京都 ‹kyō-to› "Kyoto" in a test sentence, its **W** can be decoded as "S|B-2|B-3|…" to represent that it has been a prefix for a bi-character word 東京 ‹tō-kyō› "Tokyo" and a tri-character one 東京都 ‹tō-kyō-to› "Tokyo Metropolis." If it happens to also match a word longer than five characters, the decoded form may be "S|B-2|B-3|…|B-+|…" with "B-+" denoting the relationship. Based on Figure 3, it is somewhat difficult to decide whether grouping **C** with **W** or not. Preliminary experiments realize that, in exchange for a better finalization of fully trained models, a **WC** concatenated vector like "S|B-2|B-3|…|東" should apply in the example above, while giving up some potential of dynamic lexicon expansion. Please take a sneak peek of Table 7 and a glimpse of Table 8, as they are eventually affected, and kindly note that using standalone **W** may increase $F_1$-After to 99.6 at most, but decreases $F_1$-Before to 97.5 or even lower.

### 3.4 Final Design

With all the additional features on the above, a 22-class labeling scheme is then designed empirically with combinations of $\{B, 2, 3, I, E, S\} \times \{+, \varepsilon\} \times \{F, \varepsilon\}$, tag sets for B23IES, word length, and character type, respectively. The word length class set marks a character of a six-or-more-character word as "+," otherwise blank. The character types are categorized as either Japanese letters or not. "F" covers the latter by including both functional symbols like punctuations and foreign alphanumeric characters. Labels for Japanese letters remain intact.

The templates that diversify species of feature conjunctions within the 3 families **C**, **LC**, and **WC**, are shared generating functions as previous subsection applied:

- 5 unigrams: $X_i(\cdot)$ for $-2 \leq i \leq 2$
- 4 bigrams: $X_i X_{i+1}(\cdot)$ for $-2 \leq i \leq 1$
- 3 skip-grams: $X_{i-1} X_{i+1}(\cdot)$ for $-1 \leq i \leq 1$
- 3 trigrams: $X_i X_{i+1} X_{i+2}(\cdot)$ for $-2 \leq i \leq 0$

Consequently, there are $3 \times (5 + 4 + 3 + 3)$ resultant species. Those binary features are then factorized with Recall-biased scales indicating each species' average adaptive capability. A linear-chain CRFs model thereby estimates individual's strength to a struggle of 22 classes.

## 4 Adaptation by Dynamic Lexicon Expansion

This section describes a somewhat unrealistic scenario: oracle tests without OOV issues. The purpose is merely to demonstrate the ability of hot-swapping lexicon. For every test sample, acquiring **LC**/**WC** related traits to mutate building blocks of binary feature species could lightly alter the behavior the constant model and slightly improve the accuracy. Table 8 presents the experiment results of BCCWJ variations. To get a sense about how good an industrial-grade system can be, while accommodating different word segmentation regulations, the test set of SUW-NT are also used to evaluate MeCab[9] 0.996 with UniDic[10] 2.1.2, since this latest UniDic is highly likely including all the words from BCCWJ and more. Additionally, KyTea[11] 0.46 and 0.47 are evaluated with SSUW-NT test set, because 0.47 might have been updated significantly with partially annotated data from 0.46. Results are listed in Table 8. Based on the note from the section of **WC**, $F_1$ scores from the both tables may imply that 98.5 could be a reasonable baseline, in the mean time, ≥99.6 would be hard to attain if model retraining or active learning is infeasible.

---

[9] http://taku910.github.io/mecab/
[10] https://osdn.net/projects/unidic/

| Script Form | Segmentation (* marks wrong predictions) | Alternative Script Form ‹Transliteration› "Translation" | Reference |
|---|---|---|---|
| 1 | エルマー｜と｜りゅう<br>*エルマー｜とりゅう | エルマー｜と｜竜<br>‹e-ru-mā to ryū›<br>"Elmer and (the) dragon" | (Kitagawa and Komachi, 2017) |
| 1 | キ｜ニ｜ナリ｜マス<br>キ｜ニ｜ナ｜リ｜マ｜ス<br>*キニナリマス | 気｜に｜なり｜ます<br>‹ki ni na-ri ma-su›<br>"concerning" | (Kaji and Kitsuregawa, 2014) |
| 1 & 2 | おやつ｜たーいむ<br>お｜や｜つ｜たーいむ<br>*おやつ｜たー｜い｜む | お八つ｜タイム<br>‹o-ya-tsu ta-i-mu›<br>"(prefix for polite) tea time" | |

Table 9: Selected Error Cases from Previous Works

## 5 Error Analysis

Table 9 collects related works' error cases (Kaji and Kitsuregawa, 2014; Kaji et al., 2015; Kitagawa and Komachi, 2016; Kitagawa and Komachi, 2017) that are also segmented inappropriately by the proposed method. At least 14 previously reported errors are correctly predicted and therefore omitted here. Inconsistent segmentation standards are listed first and then converted to SSUW without loss of generality. False predictions from this work's system are all related to alternate forms of scripts, e.g., (1) replacing kanji/hiragana with hiragana/katakana and (2) spelling loan words differently. The system makes no mistakes if those cases were in lexically normalized forms. It is probably worth mentioning that りゅう vs. 竜 ‹ryū› "dragon" in the first case is commonly interchangeable, and the former one is usually preferred (as it is the officially translated title of a children's novel), unless it is forming 竜王 ‹ryū ō› "dragon king" or any similar multi-kanji word. Sometimes a non-kanji usage also implies western dragons rather than Asian ones. Furthermore, only $X_{-1}X_0X_1(\mathbf{C})$ can predict the first case right. Such phenomenon suggests that finer grained clusters of feature conjunctions, e.g. binning as previously mentioned, could be beneficial.

## 6 Conclusion and Perspectives

This work investigates character-based feature engineering in detail for Japanese word segmentation. New features and labeling scheme have been carefully designed to both enhance the linear-chain CRFs model performance and enable dynamic lexicon expansion, with respect to the correlation analysis done by Goodman and Kruskal's tau. Experiments show that not only the proposed method reaches competitive performance in terms of $F_1$ score on BCCWJ corpus, but also establishes rationales of feature scaling tricks based on each conjunctive feature's predictive power for either IV or OOV words.

Error analysis reveals that character types may introduce bias to implicit word formation rules, unless lexical normalization has been properly executed to accommodate conventions of when to use which script: hiragana, katakana, or kanji. In other words, it would be an interesting future work of integrating a unified representation of script types. Meanwhile, more corpora, machine learning models, and evaluation metrics, should be tested, in order to further estimate how robust the proposed feature engineering techniques can be, especially for lexicon expansion. The scope could even be expanded (back) to Chinese word segmentation. Finally, moving toward to the next stage of part-of-speech tagging is in the plan.

---
[11] http://www.phontron.com/kytea/